\pdfoutput=1

\documentclass[11pt,a4paper]{article}
\usepackage[hyperref]{acl2021}
\usepackage{times}
\usepackage{latexsym}
\usepackage{graphicx}
\usepackage{booktabs}
\usepackage{xcolor}
\usepackage{tabularx}
\usepackage{enumitem}

\usepackage{microtype}

\aclfinalcopy % Uncomment this line for the final submission
\def\aclpaperid{689} %  Enter the acl Paper ID here

\definecolor{bluep}{rgb}{0.2, 0.2, 0.6}

\title{Ruddit: Norms of Offensiveness for English Reddit Comments}

\author{Rishav Hada$^{1,}$\thanks{\hspace{1.5mm}Both authors contributed equally.}  , Sohi Sudhir$^{1,\color{bluep}{*}}$, Pushkar Mishra$^{2}$, Helen Yannakoudakis$^{3}$, \\ \textbf{Saif M. Mohammad$^{4}$, Ekaterina Shutova$^{1}$} \\ $^{1}$ILLC, University of Amsterdam \\ $^{2}$Facebook AI, London \\ $^{3}$Dept. of Informatics, King's College London \\ $^{4}$National Research Council Canada \\ \small{\texttt{rishavhada@gmail.com, sohigre@gmail.com, pushkarmishra@fb.com,}} \\\small{\texttt{helen.yannakoudakis@kcl.ac.uk, saif.mohammad@nrc-cnrc.gc.ca, e.shutova@uva.nl}}} 
\date{}

\begin{document}

\maketitle

\begin{abstract}

\textit{\textbf{Warning:} This paper contains comments that may be offensive or upsetting.}

On social media platforms, hateful and offensive language negatively impact the mental well-being of users and the participation of people from diverse backgrounds. Automatic methods to detect offensive language have largely relied on datasets with categorical labels. However, comments can vary in their degree of offensiveness. We create the first dataset of English language Reddit comments that has \textit{fine-grained, real-valued scores} between -1 (maximally supportive) and 1 (maximally offensive). The dataset was annotated using  \emph{Best--Worst Scaling}, a form of comparative annotation that has been shown to alleviate known biases of using rating scales. We show that the method produces highly reliable offensiveness scores. Finally, we evaluate the ability of widely-used neural models to predict offensiveness scores on this new dataset.
\end{abstract}

\vspace*{-3mm}
\section{Introduction}
\label{sec:intro}
\setitemize[0]{leftmargin=*}
\setenumerate[0]{leftmargin=*}

Social media platforms serve as a medium for exchange of ideas on a range of topics, from the personal to the political. 
This exchange can, however, be disrupted by offensive or hateful language. Such language is pervasive online \citep{fbabuses}, and exposure to it may have numerous negative consequences for the victim's mental health \citep{Munro}. Automated offensive language detection has thus been gaining interest in the NLP community, as a promising direction to better understand the nature and spread of such content. 

There are several challenges 
in the automatic detection of offensive language \citep{prob}. 
The NLP community has adopted various definitions for offensive language, classifying it into specific categories. For example, \citet{waseem-hovy} classified comments as \textit{racist, sexist, neither}; \citet{davidson} as \textit{hate-speech, offensive but not hate-speech, neither offensive nor hate-speech} and \citet{founta} as \textit{abusive, hateful, normal, spam}.  \citet{schmidt-wiegand-2017-survey, Fortuna2018ASO, survey, ethics} summarize the different definitions. However, these categories 
have significant overlaps with each other, creating ill-defined boundaries, thus introducing ambiguity and annotation inconsistency \citep{founta}. A further challenge is that after encountering several highly offensive comments, an annotator might find subsequent moderately offensive comments to not be offensive ({\it de-sensitization}) \citep{kurrek-etal-2020-towards, Soral2018ExposureTH}.

At the same time, existing approaches do not take into account that comments can be offensive to a different degree. Knowing the degree of offensiveness of a comment has practical implications, when taking action against inappropriate behaviour online, as it allows for a more fine-grained analysis and prioritization in moderation.

The representation of the offensive class in a dataset is often boosted using different strategies. The most common strategy used 
is key-word based sampling. This results in datasets that are rich in explicit offensive language
(language that is unambiguous in its potential to be offensive, such as those using slurs or swear words \citep{waseem-etal-2017-understanding})
but lack cases of implicit offensive language 
(language with its true offensive nature obscured due to lack of unambiguous swear words, usage of sarcasm or offensive analogies, and others \citep{waseem-etal-2017-understanding, wiegand-etal-2021-implicitly-abusive})
\citep{waseem, weigand_bias}. 
key-word based sampling often results in spurious correlations (e.g., sports-related expressions such as \textit{announcer} and {\it sport} occur very frequently in offensive tweets).
Lastly, existing datasets consider offensive comments in isolation from the wider conversation of which they are a part. Offensive language is, however, inherently a social phenomenon and its analysis has much to gain from taking the conversational context into account \citep{gao}. 

In this paper, we present the first dataset of 6000 English language Reddit comments that has \textit{fine-grained, real-valued scores} between -1 (maximally supportive) and 1 (maximally offensive) 
-- normative offensiveness ratings for the comments. 
For the first time, we use comparative annotations to detect offensive language.
In its simplest form, comparative annotations involve giving the annotators two instances at a time, and asking which exhibits the property of interest to a greater extent. This alleviates several annotation biases present in standard rating scales, such as scale-region bias \citep{presser2004questions, asaadi-etal-2019-big}, and improves annotation consistency \cite{kiritchenko-mohammad-2017-best}. However, instead of needing to annotate $N$ instances, one now needs to annotate $N^2$ instance pairs---which can be prohibitive. Thus, we annotate our dataset using an efficient form of comparative annotation called \emph{Best--Worst Scaling (BWS)} \cite{bws_jj, louviere_flynn_marley_2015, kiritchenko-mohammad-2016-capturing, kiritchenko-mohammad-2017-best}. 

By eliminating different offensiveness categories, treating offensiveness as a continuous dimension, and eliciting comparative judgments from the annotators (based on their understanding of what is offensive), we alleviate the issues regarding category definitions and arbitrary category boundaries discussed earlier. By obtaining real-valued offensiveness scores, different thresholds can be used in downstream applications to handle varying degrees of offensiveness appropriately. By framing the task as a comparative annotation task, we obtain consistent and reliable annotations. We also greatly mitigate issues of annotator de-sensitization as one will still be able to recognize if one comment is more offensive than another, even if they think both comments are not that offensive.

In contrast to existing resources, which provide annotations for individual comments, our dataset includes conversational context for each comment (i.e. the Reddit thread in which the comment occurred). We conduct quantitative and qualitative analyses of the dataset to obtain insights into how emotions, identity terms, swear words, are related to offensiveness. Finally, we benchmark several widely-used neural models in their ability to predict offensiveness scores on this new dataset.\footnote{Dataset and code available at:\\ \url{https://github.com/hadarishav/Ruddit}.}

\section{Related Work}
\subsection{Offensive Language Datasets} 
\label{existingdatasets}

Surveys by \citet{schmidt-wiegand-2017-survey, Fortuna2018ASO, survey, vidgen2020directions} discuss 
various existing datasets and their compositions in detail. 
\citet{waseem-hovy, davidson, founta} created datasets based on Twitter data. Due to prevalence of the non-offensive class in naturally-occurring data \citep{waseem, founta}, the authors devised techniques to boost the presence of the offensive class in the dataset. \citet{waseem-hovy} used terms frequently occurring in offensive tweets, while \citet{davidson} used a list of hate-related terms to extract offensive tweets from the Twitter search API. \citet{park_bias}, \citet{weigand_bias}, and \citet{davidson_bias} show that the \citet{waseem-hovy} dataset exhibits topic bias and author bias due to the employed sampling strategy. 
\citet{founta} boosted the representation of offensive class in their dataset by analysing the sentiment of the tweets and checking for the presence of offensive terms. In our work, we employ a hybrid approach, selecting our data in three ways: specific topics, emotion-related key-words, and random sampling.

Past work has partitioned offensive comments into {\it explicitly offensive} (those that include profanity---swear words, taboo words, or hate terms)  and \emph{implicitly offensive} (those that do not include profanity) \cite{waseem-etal-2017-understanding, caselli-etal-2020-feel, wiegand-etal-2021-implicitly-abusive}. 
Some other past work has defined explicitly and implicitly offensive instances a little differently: \citet{sap2020socialbiasframes} considered factors such as obviousness, intent to offend and biased implications, \citet{breitfeller-etal-2019-finding} considered factors such as the context and the person annotating the instance, and \citet{razo-kubler-2020-investigating} considered the kind of lexicon used. 
Regardless of the exact definition, implicit offensive language, due to a lack of lexical cues, is harder to classify not only for computational models, but also for humans. 
In our work, we consider implicitly offensive comments as those offensive comments that do not contain any swear words.

\citet{exmachina,exmachina-2} created three different datasets from Wikipedia Talk pages, focusing on aggression, personal attacks and toxicity. The comments were sampled at random from a large dump of English Wikipedia, and boosted by including comments from blocked users. For the personal attacks dataset, \citet{exmachina} used two different kinds of labels: ED (empirical distribution), OH (one hot). In case of ED, the comments were assigned real-valued scores between 0 and 1 representing the fraction of annotators who considered the comment a personal attack. While these labels were introduced to create a separation between the nature of comments with a score of $1.0$ and those with a score of $0.6$ (which would otherwise be classified as attacks),
they are discrete. In our work, using the BWS comparative annotation setup, we assign fine-grained continuous scores to  comments to denote their degree of offensiveness.

\subsection{Best--Worst Scaling (BWS)}
\label{BWS}
BWS was proposed by \citet{bws_jj}. \citet{kiritchenko-mohammad-2017-best} have experimentally shown that BWS produces more reliable fine-grained scores than the scores acquired utilizing rating scales. In the BWS annotation setup, the annotators are given an n-tuple (where n $>$ 1, and  commonly $n = 4$), and asked which item is the best and which is the worst (best and worst correspond to the highest and the lowest with respect to a property of interest). Best--worst annotations are particularly efficient when using 4-tuples, as each annotation results in inequalities for 5 of the 6 item pairs. For example, a 4-tuple with items A, B, C, and D, where A is the best, and D is the worst, results in inequalities: A$>$B, A$>$C, A$>$D, B$>$D, and C$>$D. 
Real-valued scores of associations are calculated
between the items and the property of interest from the best--worst annotations for a set of 4-tuples \citep{omre, RePEc:elg:eechap:14820_8}. The scores can be used to rank items by the degree of association with the property of interest.  Within the NLP community, BWS has thus far been used only for creating datasets for relational similarity \citep{rs}, word-sense disambiguation \citep{jurgens-2013-embracing},  word–sentiment  intensity \citep{ws}, phrase  sentiment composition \citep{kiritchenko-mohammad-2016-capturing}, and tweet-emotion intensity \citep{mohammad-bravo-marquez-2017-emotion, mohammad-kiritchenko-2018-understanding}. 
Using BWS, we create the first dataset with degree of offensiveness scores for social media comments.

\section{Data collection and sampling}
\label{sec:emotion_calculation}

We extracted Reddit data from the Pushshift repository \citep{baumgartner2020pushshift} using \emph{Google BigQuery}. Reddit is a social news aggregation, web content rating, and discussion website. 
It contains forums called \textit{subreddits} dedicated to specific topics. Users can make a \textit{post} on the subreddit to start a discussion. Users can \textit{comment} on existing posts or comments to participate in the discussion. As users can also reply to a comment, the entire discussion has a hierarchical structure called the comment \textit{thread}.  
We divided the extracted comments into 3 categories based on their subreddit source:\\[-22pt]
\begin{enumerate}
    \item \textbf{Topics (50\%)}: Contains comments from topic-focused subreddits: \textit{AskMen, AskReddit, TwoXChromosomes, vaxxhappened, worldnews, worldpolitics}. These subreddits were 
    chosen to cover a diverse range of topics. \textit{AskReddit, vaxxhappened, worldnews, worldpolitics} 
    discuss generic themes. \textit{TwoXChromosomes} contains women's perspectives on various topics and \textit{AskMen} contains men's perspectives.
\vspace*{-2mm}    
    \item \textbf{ChangeMyView (CMV) (25\%)}: The CMV subreddit (with over a million users) has posts and comments on controversial
    topics. 
 \vspace*{-2mm}   
    \item \textbf{Random (25\%)}: Contains comments from random subreddits. 
\end{enumerate}
We selected 808 posts from the subreddits based on 
criteria such as date, thread length, and post length.
(Further details in the Appendix \ref{sec:postcriteria}.)
We took the first 25 and the last 25 comments
per post (skipping comments that had [\textsc{deleted}] or [\textsc{removed}] as comment body).
The first responses are likely to be most relevant to the post. The final comments indicate how the discussion ended.
We sampled 6000 comments from this set for annotation.

The goal of the sampling was to increase the proportion of offensive and emotional comments.
Emotions are highly representative of one's mental state, which in turn are associated with their behaviour \citep{poria2019emotion}. For example, \citet{Jay2008ThePO} show that people tend to swear when they are angry, frustrated or anxious. 

Studies have shown that the primary dimensions of emotion are valence, arousal, and dominance (VAD) \citep{osgood, Russell1980, russell2003}. \textbf{\textit{Valence}} is the positive -- negative or pleasure--displeasure dimension.\textbf{\textit{ Arousal}} is the excited--calm or active--passive dimension. \textbf{\textit{Dominance}} is powerful--weak or `have full control'--`have no control' dimension \citep{vad}. To boost the representation of offensive and emotional comments in our dataset, we up-sampled comments that included low-valence (highly negative) words and those that included high-arousal words (as per the NRC VAD lexicon \citep{vad}).\footnote{In some initial pilot experiments, we found this approach of sampling low valence and high arousal comments to result in a greater number of offensive comments.} The manually constructed NRC VAD lexicon includes 20,000 English words, each with a real-valued score between 0 and 1 in the V, A, D dimensions. 

In order to do this upsampling, we first defined the valence score of each comment as the average valence score of the negative words within the comment
(A negative word is defined as a word with a valence score $ \leq 0.25$ in the VAD lexicon.) Similarly, we defined the arousal score for a comment as the average arousal score of high-arousal words in each comment.
(A high-arousal word is defined as a word with an arousal score $ \geq 0.75$.)
We selected comments from the comment pool such that 50\% 
 
were from the \textit{Topics} category, 25\% 
 
from the \textit{CMV} category, and 25\% 

from the \textit{Random} category. Within each category, 33\% of the comments were those that had the lowest valence scores, 33\% of the comments were those that had the highest arousal scores, and the remaining were chosen at random.

\section{Annotation}
\label{sec:data-annotation}

The perception of `offensiveness' of a comment can vary from person to person. Therefore, we used crowdsourcing to annotate our data. Crowdsourcing helps us get an aggregation of varied perspectives rather than expert opinions which can leave out offensiveness in a comment that lies outside the `typical’ offensiveness norms \citep{blackwell}. 
We carried out all the annotation tasks on Amazon Mechanical Turk (AMT). Due to the strong language, an adult content warning was issued for the task. Reddit is most popular in the US, which accounts for ~50\% of its desktop traffic \cite{reddit-demo}. Therefore, we restricted annotators to those residing in the US. To maintain the quality of annotations, only annotators with high approval rate were allowed to participate. 

\subsection{Annotation with Best--Worst Scaling}
We followed the procedure described in \citet{kiritchenko-mohammad-2016-capturing} to obtain BWS annotations. Annotators were presented with 4 comments (4-tuple) at a time and asked to select the comment that is most offensive (least supportive) and the comment that is least offensive (most supportive). We randomly generated 2N distinct 4-tuples (where N is the number of comments in the dataset), such that each comment was seen in eight different 4-tuples and no two 4-tuples had more than 2 items in common. We used the script provided by \citet{kiritchenko-mohammad-2016-capturing} to obtain the 4-tuples to be annotated.\footnote{\url{http://saifmohammad.com/WebPages/BestWorst.html}}

\citet{kiritchenko-mohammad-2016-capturing} show that in a word-level sentiment task, using just three annotations per 4-tuple produces highly reliable results. However, since we work with long comments and 
a relatively more difficult task, 
we got each tuple annotated by 6 annotators. Since each comment is seen in 8 different 4-tuples, we obtain 8 X 6 = 48 judgements per comment. \label{bws_score}

\begin{table*}[t!]
\centering
{\small
\begin{tabular}{llllll}
\toprule
\textbf{\# Comments} &  \textbf{\# Annotations per Tuple} & \textbf{\# Annotations} & \textbf{\# Annotators} & \textbf{SHR Pearson}        & \textbf{SHR Spearman}        \\ \midrule

6000       & 6                                                                  & 95,255        & 725          & 0.8818 $\pm$ 0.0023 & 0.8612 $\pm$ 0.0029 \\ \bottomrule
\end{tabular}%
}
\caption{Ruddit annotation statistics and split-half reliability (SHR) scores.}
\label{tab:shr}

\end{table*}

\subsection{Annotation Task and Process}
\label{sec:annotations}

In our instructions to the annotators, we defined offensive language as comments that include but are not limited to [being hurtful (with or without the usage of abusive words)/ being intentionally harmful/ treating someone improperly/ harming the `self-concept' of another person/ aggressive outbursts/ name calling/ showing anger and hostility/ bullying/ hurtful sarcasm]. We also encouraged the annotators to follow their instincts. By framing the task in terms of comparisons and providing a broad definition of offensiveness, we avoided introducing artificial categories and elicit responses guided by their intuition of the language.

Detailed annotation instructions are made publicly available (Figure \ref{fig:instructions} in Appendix \ref{sec:qa}).\footnote{AMT task interface with instructions: \url{https://hadarishav.github.io/Ruddit/}} A sample questionnaire
is shown in Figure \ref{fig:questionnaire} in 
Appendix \ref{sec:qa}. For quality control purposes, we manually annotated around 5\% of the data ourselves beforehand. We will refer to these instances as \textit{gold questions}. The gold questions were interspersed with the other questions. 
If a worker's accuracy on the gold questions fell below 70\%, they were refused further annotation and all of their annotations were discarded. The discarded annotations were published again for re-annotation. We received a total of 95,255 annotations by 725 crowd workers. 

The BWS responses were converted to scores using a simple counting procedure \citep{omre, RePEc:elg:eechap:14820_8}. For each item, the score is the proportion of times the item is chosen as the most offensive minus the proportion of times the item is chosen as the least offensive. We release the aggregated annotations as well as the individual annotations of Ruddit, to allow further work on examining and understanding the variability.\footnote{We provide the comment IDs and not the comment body, in accordance to the GDPR regulations. Comment body can be extracted using the Reddit API.}

\begin{table*}[t!]
\centering
{\small
\begin{tabularx}{\textwidth}{lXl}
\toprule
\textbf{Bin} & \textbf{Comment}                                                                                                                 & \textbf{Score} \\ \midrule
1            & Don't worry, she's going to be fine.                                                                                             & $-0.75$        \\
             & I see you too are a man of culture;)                                                                                             & $-0.604$       \\ [2pt] 
2            & This is so sexy! Love it!                            & $-0.562$        \\
             & ``I live with my ex, but it's totally cool, we're just friends"                                                                   & $-0.229$       \\ [2pt] 
3            & Not sure why Im being down voted? Why does the truth bother so many people?                                                      & $-0.191$       \\
 &
  I presented a hypothetical question to you. I did not even claim that you made that argument. Unfortunately that is not a straw man. So, care to answer that question again? &
  0.083 \\[2pt]
4            & {Don't forget Vaccines cause autism. And torture is awesome.  We should murder the families of terrorists. }      & 0.5           \\

             & \textbf{What is your angle, Kim?? Is this some Hitler BS where you sign a peace treaty and then start WWIII? Or did you finally just grow a brain? Because neither sound particularly more likely than the other... }           & 0.521            \\[2pt] 
5            & If you support trump kill yourself, painfully                                                                           & 0.604          \\

             & shut the fuck up bitch. It’s Bernie or Bust  nobody is voting for Biden, now get the fuck out of here you cunt     & 0.958          \\ \bottomrule
\end{tabularx}
}
\caption{Sample comments from Ruddit for each of the 5 score bins. Comment in bold is implicitly offensive.}
\label{tab:comments}
\end{table*}

\subsection{Annotation Reliability}
We cannot use standard inter-annotator agreement measures to ascertain the quality of comparative annotations. The disagreement that arises in tuples having two items that are close together in their degree of offensiveness is a useful signal for BWS (helping it give similar scores to the two items).
The quality of annotations can be measured by measuring the reproducibility of the end result -- if repeated manual annotations from multiple annotators can produce similar rankings and scores, then, one can be confident about the quality of annotations received. To assess this reproducibility, we computed average \emph{split-half reliability} (SHR) values over 100 trials. SHR is a commonly used approach to determine consistency in psychological studies. 

For computing SHR values, the annotations for each 4-tuple were randomly split in two halves. Using these two splits, two sets of rankings were determined. We then calculated the correlation values between these two sets. This procedure was repeated 100 times and the correlations were averaged. A high correlation value indicates that the annotations are of good quality. 
Table \ref{tab:shr} shows the SHR for our annotations.
SHR scores of over 0.8 indicate substantial reliability.

\section{Data Analysis}
\label{sec:data-analysis}

In this section, we analyze various aspects of the data, including: the distribution of scores, the association with identity terms, the relationship with emotion dimensions, the relationship with data source, and the role of swear words.

\begin{figure}[t!]
\centering
\resizebox{\columnwidth}{!}{%
  \includegraphics[scale=0.3]{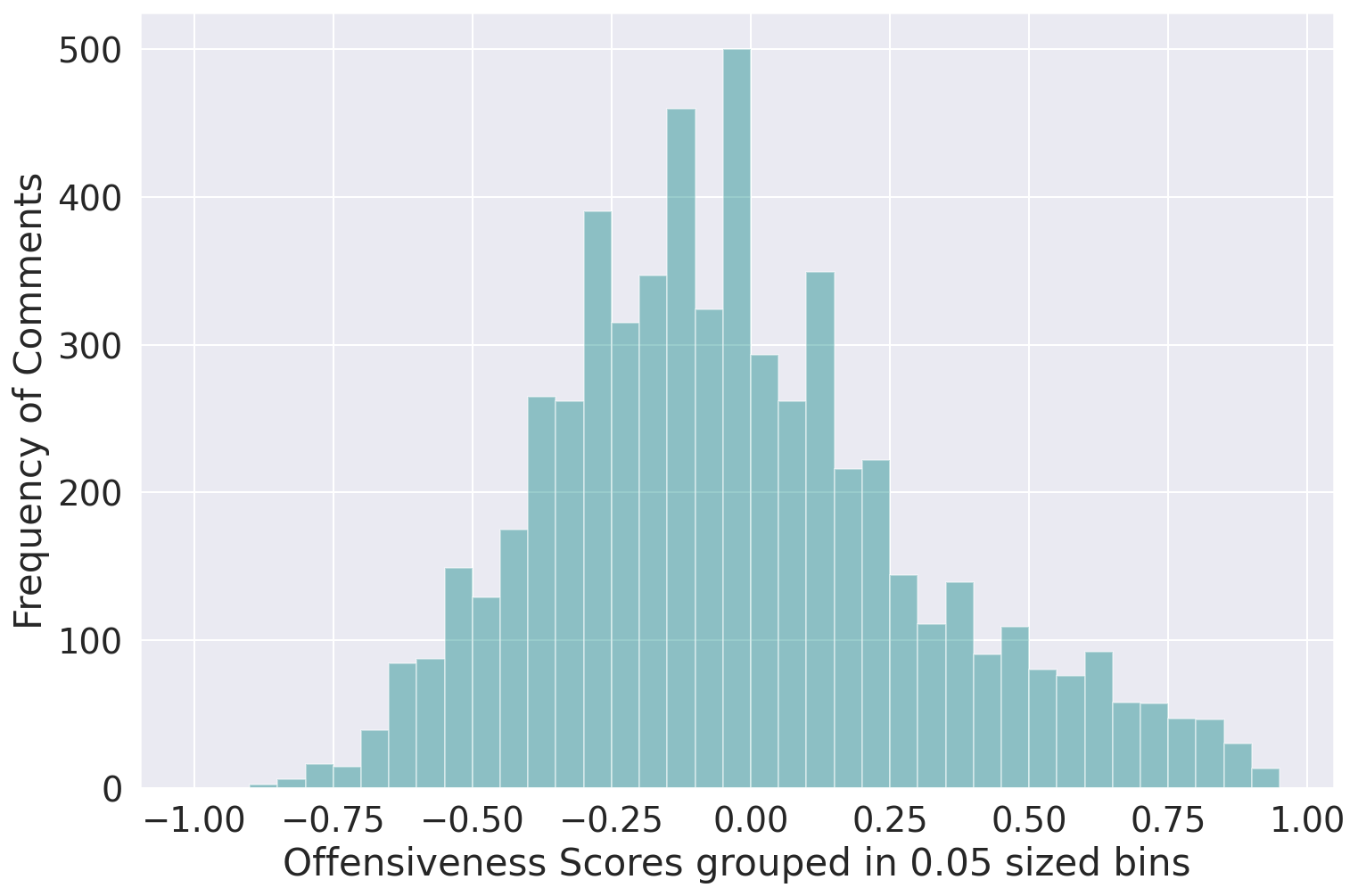}}
  \caption{A histogram of frequency of comments--degree of offensiveness. Degree of offensiveness scores are grouped in bins of size $0.05$.}
  \label{fig:our_data}
 % \vspace*{-3mm}
\end{figure}

\paragraph{Distribution of Scores}
Figure \ref{fig:our_data} shows a histogram of frequency of comments vs.\@ degree of offensiveness, over 40 equi-spaced score bins of size $0.05$. We observe a normal distribution. 

To analyze the data, we placed the comments in 5 equi-spaced score bins of size $0.4$
(bin 1: $-1.0$ to $-0.6$, bin 2: $-0.6$ to $-0.2$, and so on). 
Table \ref{tab:comments} shows some comments from the dataset (more examples can be found in Appendix \ref{sec:sampledata} Table \ref{tab:comment-examples}). 
We observed that bin 1 primarily contains supportive comments while bin 2 shows a transition from supportive to neutral comments. Bin 3 is dominated by neutral comments but as the score increases the comments become potentially offensive and bins 4 \& 5 predominantly contain offensive comments. 
It is interesting to note that 
bin 4 contains
some instances of implicit offensive language such as \textit{`You look like a lesbian mechanic who has a shell collection'}. In their paper, \citet{wiegand-etal-2021-implicitly-abusive} explore the category of such ``\textit{implicity abusive comparisons}'', in depth. More examples of implicitly offensive comments present in our dataset can be found in table \ref{tab:comments} and table \ref{tab:comment-examples} (in Appendix \ref{sec:sampledata}).

To explore whether specific bins capture specific topics or key-words, we calculated Pointwise Mutual Information (PMI) scores of all the unique words in the comments (excluding stop words) with the five score bins. 
Table \ref{tab:pmi} shows the top scoring words for each bin. We observed that bins 1, 2, and 3 exhibit a strong association with supportive or neutral words, while bins 4 and 5 show a strong association with swear words and identity terms commonly found in offensive contexts. 

\begin{table}[t!]
\centering
{\small
\begin{tabularx}{\columnwidth}{lX}
\toprule
\textbf{Bin} & \textbf{Words}                  \\ \midrule
1            & awesome, thanks, appreciate     \\
2            & songs, headphones, sweet, movie \\
3            & gap, sacrifice, employee         \\
4            & muslim, fucked, gay, ass, raped \\
5            & dick, fuck, asshole, ass, shut \\ \bottomrule
\end{tabularx}%
}
\caption{Top PMI scoring words for each of the 5 offensiveness-score bins. Degree  of  offensiveness scores are grouped in bins of size $0.4$.}
\label{tab:pmi}
\end{table}

\paragraph{Identity terms}
A common criticism of the existing offensive language datasets is that in those datasets, certain identity terms (particularly those referring to minority groups) occur mainly in texts that are offensive \citep{sap-etal-2019-risk, davidson_bias, weigand_bias, park_bias, dixon_bias}.  This leads to high association of targeted minority groups (such as Muslims, females, black people and others) with the offensive class(es). 
This bias, in turn, is captured by the computational models trained on such datasets. 
As mentioned earlier, in Ruddit, certain words such as  \textit{gay, trans, male, female, black, white} were found to exhibit a relatively higher association with the offensive bins than with the supportive bins. In order to probe the effect of this on the computational models, we created a variant of Ruddit by replacing all the identity terms (from the list given in Appendix \ref{sec:identity}) in the comments with the [\textit{group}] token and observed the effect on the models' performance.  We refer to this variant of the dataset as the \textbf{\textit{identity-agnostic dataset}}. We analyse the models' performance in the next section.

\paragraph{Offensiveness vs.\@ emotion} 

As discussed earlier, our emotions impact the words we use in text. We examined this relationship quantitatively using Ruddit and the NRC VAD Lexicon (which has intensity scores along the valence, arousal, and dominance dimensions). 

For each comment in Ruddit, we calculated three scores that captured the intensities of the V, A, D words (the averages of the intensities of the V/A/D words in the comment), using the entire lexicon.

We then determined the correlation between each of the three scores and the degree of offensiveness. Only comments containing at least 4 words from the VAD lexicon were considered for the score and correlation calculation. A total of 4831 comments qualified the criteria.
See Table \ref{tab:emotions}.

\begin{figure}[t!]
\includegraphics[width=\columnwidth, height = 5cm]{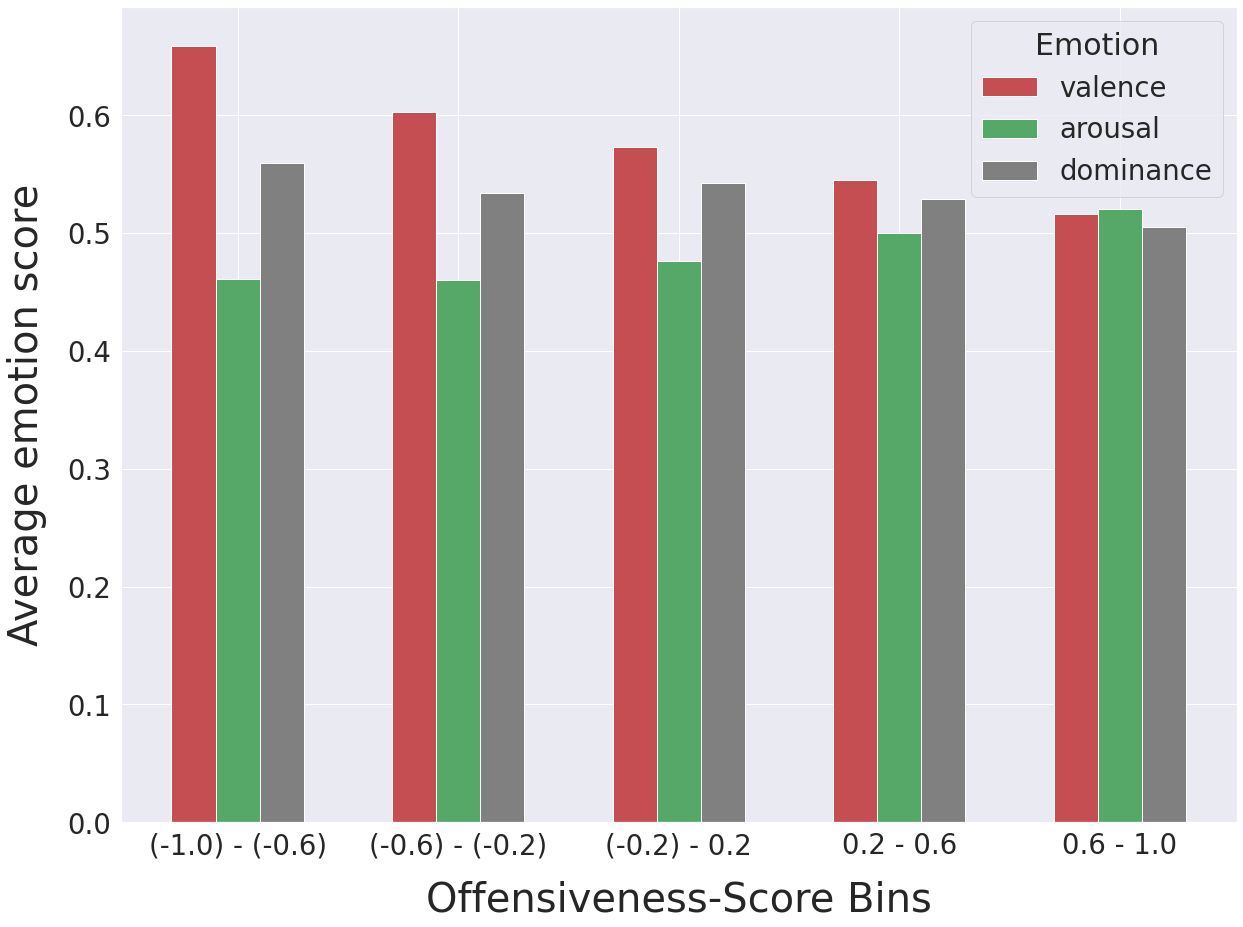}
\centering
\caption{Average Valence, Arousal, and Dominance scores of comments in various offensiveness-score bins.
}

\label{fig:vad_emoscore}
\end{figure}

From the table, we can observe that \textit{valence} is weakly inversely correlated, \textit{arousal} is weakly correlated, and \textit{dominance} does not exhibit notable correlation with offensiveness. This behaviour can also be observed in Figure \ref{fig:vad_emoscore} that shows a plot of the average V, A, and D scores of comments in the five equi-spaced offensiveness-score bins.

Note the clear trend that as we look at bins with more offensive comments, the average valence of the comments decreases and the average arousal increases.

\begin{table}[t!]
\centering
{
\begin{tabular}{lr}
\toprule
\textbf{Emotion} & \textbf{Pearson's r} \\ \midrule
Valence          & \textbf{$-0.301$}                           \\
Arousal          & \textbf{$0.256$}                           \\
Dominance        & $-0.086$            

\\ \bottomrule
\end{tabular}%
}
\caption{Pearson correlation values between the offensiveness scores and the emotion dimension scores.}
\label{tab:emotions}
\end{table}

\begin{figure}[t!]
% \centering
  \includegraphics[width = \columnwidth, height = 5cm]{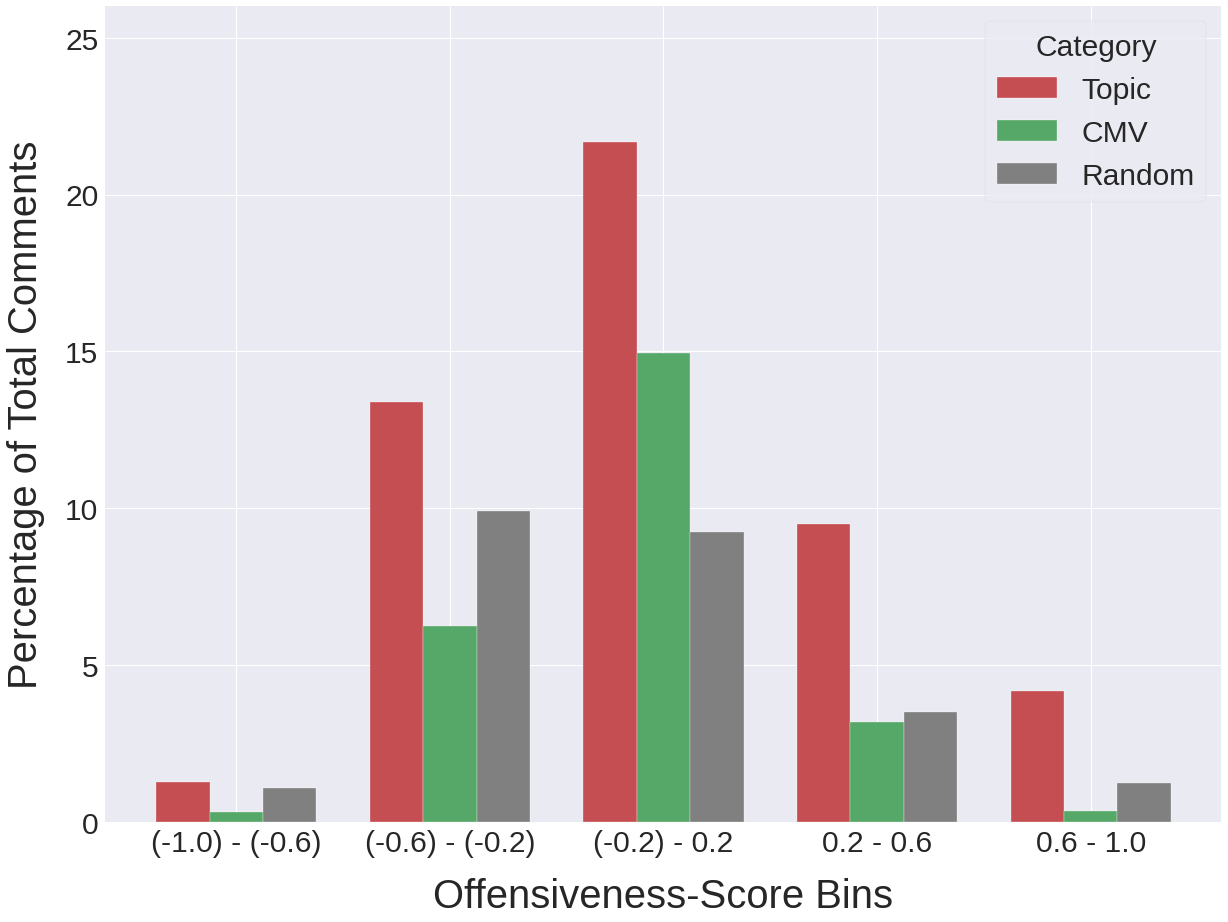}
  \centering
  \caption{Distribution of comments from each comment category in each offensiveness-score bin.}
  \label{fig:category_dist1}
%  \vspace*{-3mm}
\end{figure}

\paragraph{Offensiveness vs.\@ data source}
As mentioned earlier, comments in our dataset come from three different sources - Topics, CMV, and Random. Figure \ref{fig:category_dist1} shows the distribution of comments from each source over the score bins. We observed that comments from Topics have near equal representation on both sides of the scale, while for the other two sources, comments are more prevalent in the supportive bins. The higher representation of comments from Topics than the other two sources in the offensive bins, is likely due to the fact that the Topics category includes subreddits such as \textit{worldnews} and \textit{worldpolitics}. Discussions on these subreddits covers controversial topics and lead to the usage of offensive language. We observed that \textit{worldnews} and \textit{worldpolitics} indeed have high representation in the offensive bins (Figure \ref{fig:topic_dist} in Appendix \ref{sec:identity}).

\paragraph{Swear words}

We identified 868 comments in our dataset that contain at least one
swear word from the cursing lexicon \citep{wenbo_cursing}. Comments containing swear words can have a wide range of offensiveness scores. 
To visualize the distribution, we plot a histogram of the comments containing swear words vs.\@ degree of offensiveness (see Figure \ref{fig:swear_dist} in Appendix \ref{sec:identity}).
The distribution is skewed towards the offensive end of the scale. An interesting observation is that some comments with low offensiveness scores contain phrases using swear words to express enthusiasm or to lay more emphasis, for example `\textit{Hell yes}', `\textit{sure as hell love it}', `\textit{uncomfortable as shit}' and others.
To study the impact of comments containing swear words on computational models, we created another variant of Ruddit in which we removed all the comments containing at least one
swear word. We refer to this variant as the \textbf{\textit{no-swearing dataset}}. This dataset contains 5132 comments. We analyse the models' performance on this dataset in the next section.

\paragraph{Offensiveness in different score ranges}
It is possible that comments in the middle region of the scale
may be more difficult for the computational models. 
Thus, we created a subset of Ruddit containing comments with scores from $-0.5$ to $0.5$. We call this subset (of 5151 comments), the \textbf{\textit{reduced-range dataset}}. We discuss the models' performance on this dataset in the next section.

\section{Computational Modeling}

In this section, we present benchmark experiments on Ruddit and its variants by implementing some commonly used model architectures. The task of the models was to predict the offensiveness score of a given comment. We performed 5-fold cross-validation for each of the models.\footnote{Since we have a linear regression task, we created folds using \textit{sorted stratification} \citep{stratifiedregression} to ensure that the distribution of all the partitions is similar.}

\begin{table*}[t!]
\centering
{\small
\resizebox{\textwidth}{!}{%
\begin{tabular}{lcccccc}
\toprule
\textbf{Dataset} & \multicolumn{2}{c}{\textbf{HateBERT}} & \multicolumn{2}{c}{\textbf{BERT}} & \multicolumn{2}{c}{\textbf{BiLSTM}} \\
                  & $r$ & MSE   & $r$ & MSE   & $r$ & MSE   \\ \midrule
a. Ruddit              & $0.886 \pm 0.003$  & $0.025 \pm 0.001$ & $0.873 \pm 0.005$  & $0.027 \pm 0.001$ & $0.831 \pm 0.005$  & $0.035 \pm 0.001$ \\
b. \textit{Identity-agnostic} & $0.883 \pm 0.006$  & $0.025 \pm 0.001$ & $0.869 \pm 0.007$  & $0.027 \pm 0.001$ & $0.824 \pm 0.007$  & $0.036 \pm 0.001$ \\ \midrule
c. \textit{No-swearing}       & $0.808 \pm 0.013$  & $0.023 \pm 0.001$ & $0.783 \pm 0.012$  & $0.027 \pm 0.001$ & $0.704 \pm 0.014$  & $0.036 \pm 0.002$ \\ \midrule
d. \textit{Reduced-range}     & $0.781 \pm 0.014$  & $0.022 \pm 0.001$ & $0.757 \pm 0.011$   & $0.025 \pm 0.001$ & $0.659 \pm 0.008$  & $0.033 \pm 0.001$ \\ \bottomrule
\end{tabular}%
}
}
\caption{Five-fold cross-validation results of the models on Ruddit and its variants. $r$ = Pearson's R. Note: Scores for c.\@ and d.\@ are not directly comparable to scores for a.\@ and b.\@ as they involve different score ranges.}

\label{tab:results}
\vspace*{-3mm}
\end{table*}

\subsection{Models}
\label{sec:models}
\paragraph{Bidirectional LSTM}
We fed pre-trained 300 dimensional GloVe word embeddings \citep{pennington-etal-2014-glove} to a 2-layered BiLSTM to obtain a sentence representation (using a concatenation of the last hidden state from the forward and backward direction). This sentence representation was then passed to a linear layer with a \textit{tanh} activation to produce a score between $-1$ and 1. We used \textit{Mean Squared Error} (MSE) loss as the objective function, Adam with $0.001$ learning rate as the optimizer, hidden dimension of 256, batch size of 32, and a dropout of $0.5$. The model was trained for 7 epochs.  

\paragraph{BERT}
We fine-tuned BERT$_{\textit{base}}$ \citep{devlin-etal-2019-bert}. We added a regression head containing a linear layer to the pre-trained model. We used MSE loss as the objective function, batch size of 16, and learning rate of $2e-5$ (other hyperparameters same as \citep{devlin-etal-2019-bert}). We used the \textit{AdamW} optimizer with a linear learning rate scheduler with no warm up steps. The model was trained for 3 epochs. (More details in Appendix \ref{sec:gridsearch}.)

\paragraph{HateBERT}
HateBERT \cite{caselli2020hatebert} is a version of BERT pretrained for abusive language detection in English. HateBERT was trained on RAL-E, a large dataset of English language Reddit comments from communities banned for being offensive or hateful. HateBERT has been shown to outperform the general purpose BERT model on the offensive language detection task when fine-tuned on popular datasets such as OffensEval 2019 \cite{zampieri-etal-2019-semeval}, AbusEval \cite{caselli-etal-2020-feel}, and HatEval \cite{basile-etal-2019-semeval}. 

We fine-tuned HateBERT on Ruddit and its variants. The experimental setup for this model is the same as that described for the BERT model.

\subsection{Results and Analysis}
We report Pearson correlation ($r$) and MSE, averaged over all folds. The performance of the models on Ruddit and its variants 
is shown in the Table \ref{tab:results}. Note that the performance values on the \textit{no-swearing} and the \textit{reduced-range} datasets are not directly comparable to the performance values on the full Ruddit as their score range is different. We can see that on all the datasets, the HateBERT model performs the best, followed by the BERT model. Interestingly, the model performance (for all models) does not change substantially when trained on Ruddit or the \textit{identity-agnostic} dataset. This indicates that the computational models are not learning to benefit from the association of certain identity terms with a specific range of scores on the offensiveness scale.\footnote{It should be noted that since the list of identity terms and the cursing lexicon we use is not exhaustive, our conclusions are only limited to the scope of the respective lists.}

The models show a performance drop on the \textit{no-swearing} dataset, which suggests that swear words are useful indicators of offensiveness and that the comments containing them are easier to classify. Yet, the fact that the models still obtain performance of up to $0.8$ ($r$) demonstrates that they necessitate and are able to learn other types of offensiveness features. It is also worth mentioning that even if they encounter swear words in a comment, the task is not simply to label the comment as offensive but to provide a suitable score.

Finally, the models obtained the performance of up to $0.78$ ($r$) on the \textit{reduced-range} dataset, which shows that even if the comments from the extreme ends of the offensiveness scale are removed, Ruddit still presents an interesting and feasible offensiveness scoring task.

\begin{figure*}[t!]
\includegraphics[width=\textwidth, height = 5cm]{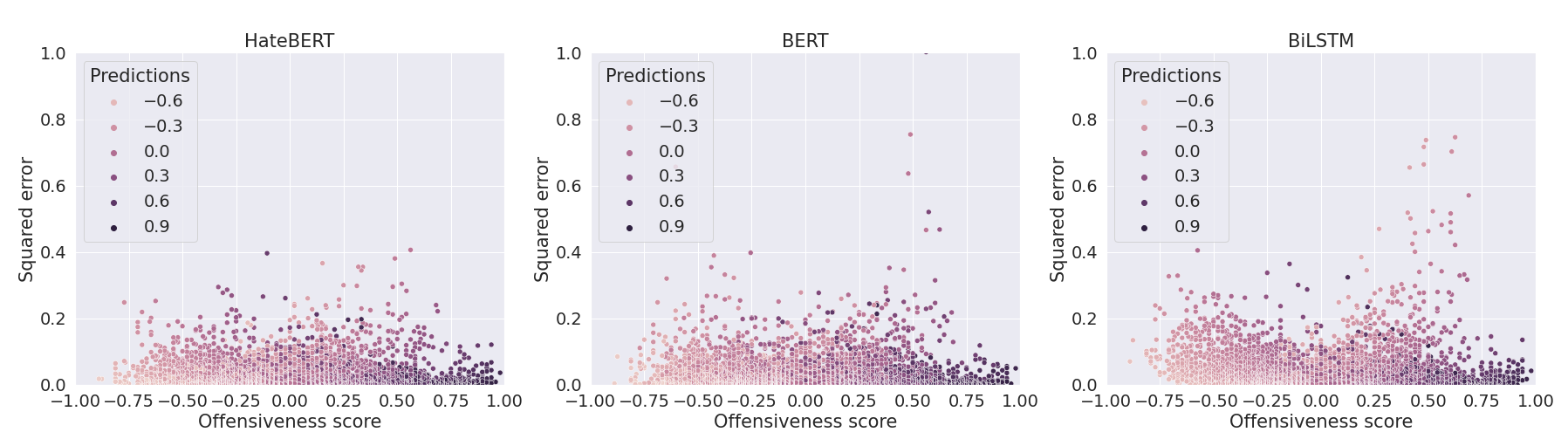}
\centering
\caption{Squared error values for the 3 models' predictions over the offensiveness score range in Ruddit.} 
  \label{fig:error}
\end{figure*}
\paragraph{Error Analysis}
Figure \ref{fig:error} shows the squared error values of the 3 models over the offensiveness score range in Ruddit. As expected, for all the models, the error in predictions is lower on both the extreme ends of the scale than in the middle region. Comments with very high or very low offensiveness scores are rich in obvious linguistic cues, making it easier for the computational models to predict scores. Most of the not-obvious, indirect implicitly offensive, and neutral comments should be present in the middle region of the offensiveness scale, making them more difficult for the models. 
It is interesting to observe that HateBERT, unlike the other two models, does not have high error values for samples within the score range $0.25$--$0.75$. 
This indicates that HateBERT is efficient in dealing with offensive language that does not lie in the extreme offensive end. BiLSTM seems relatively less accurate for samples in the supportive range ($-0.75$ to $-0.25$). This could be attributed to the less complex model architecture and the usage of GloVe word embeddings.

\section{Conclusion}
We presented the first dataset of online comments annotated for their degree of offensiveness. We used a comparative annotation technique called Best--Worst Scaling, which addresses the limitations of traditional rating scales. We showed that the ratings obtained are highly reliable (SHR Pearson $r \approx 0.88$). We performed data analysis to gain insight into the relation of emotions, data sources, identity terms, and swear words with the offensiveness scores. 
We showed that valence is inversely correlated 
with offensiveness and 
arousal is directly correlated with offensiveness.

Finally, we presented benchmark experiments to predict the offensiveness score of a comment, on our dataset. We found that computational models are not benefiting from the association of identity terms with specific range of scores on the offensiveness scale.
In future work, it would be interesting to explore the use of conversational context in computational modeling of offensiveness, as well as studying the interaction between offensiveness and emotions in more depth. We make our dataset freely available to the research community. 

\section*{Acknowledgements}
This research was funded by the Facebook Online Safety Benchmark Research award for the project ``A Benchmark and Evaluation Framework for Abusive Language Detection.”

\section*{Ethical Considerations}
\label{sec:ethics}
We create Ruddit to study, understand and explore the nature of offensive language. Any such dataset might also be used to create automatic offensive language detection systems. While we realise the importance of such systems, we also accept that any moderation of online content is a threat to free speech. Offensive language datasets or automatic systems can be misused to stifle disagreeing voices. Our intent is solely to learn more about the use of offensive language, learn about the various degrees of offensive language, explore how computational models can be enabled to watch and contain offensive language, and encourage others to do so. We follow the format provided by  \citet{bender-friedman-2018-data} to discuss the ethical considerations for our dataset.\\[-10pt]

\noindent {\bf{Institutional Review:}}
This research was funded by the Facebook Online Safety Benchmark Research award. The primary objective of this research award is the creation of publicly available benchmarks to improve online safety. This award does not directly benefit Facebook in any way. This research was reviewed by Facebook for various aspects, in particular:\\[-20pt]
\begin{itemize}
    \item Legal Review: Evaluates whether the research to be undertaken or the research performed can violate intellectual property rights. \vspace{-2mm}
    \item Policy and Ethics Review: Evaluates whether the research to be undertaken aligns with the best ethics practices. This includes several aspects such as mitigating harm to people involved, improving data privacy, and informed consent.
\end{itemize}

\noindent {\bf{Data Redistribution / User Privacy:}}
We extracted our data from the Pushshift Reddit dataset made publicly available by \citet{baumgartner2020pushshift} for research purposes.  The creators of the Pushshift Reddit dataset have provisions to delete comments from their dataset upon user’s request. We release data in a manner that is GDPR compliant. We do not provide any user-specific information. We release only the comment IDs and post IDs. Reddit’s Terms of Service do not prohibit the distribution of ids.\footnote{\url{https://www.reddit.com/wiki/api-terms}} The researchers using the dataset need to retrieve the data using the Reddit API. \\[-10pt]

\noindent {\bf{Speaker and Annotator Demographic:}}
No specific speaker demographic information is available for the comments included in Ruddit. According to the October 2020 survey published by Statista \citep{reddit-demo},  ~50\% of the Reddit's desktop traffic is from the United States. They also state that from the internet users in the US, ~21\% from ages 18-24, ~23\% from ages 25-29 and ~14\% from ages 30-49 use Reddit. 

We restricted annotators to those residing in the US. A total of 725 crowd-workers participated in the task. Apart from the country of residence, no other information is known about the annotators. The annotators are governed by AMT's privacy policy.\footnote{\url{https://www.mturk.com/help}} Pew Research Center conducted a demographic survey of AMT workers in 2016. In this survey, 3370 workers participated. They found out that 80\% of the crowd-workers on AMT are from the US \citep{amtstats}. More information about the workers who participated in their survey can be found in their article. 

It is important to include the opinions of targeted minorities and marginalized groups when dealing with the annotation of offensive language \cite{ethics, blackwell}. However, we did not  have our data annotated by the specific target demographic because it poses certain challenges. For example: identification of the target of offensive language; finding people of the target demographic group who are willing to annotate offensive language; and others.  Annotating such offensive data can be even more traumatizing for the members of the targeted minorities. Finally, Ruddit was created with the intention to look at wide ranging offensive language of various degrees as opposed to detecting offensive language towards specific target groups.\\[-10pt]

\noindent {\bf Annotation Guidelines:} We created our annotation guidelines drawing inspiration from the community standards set for offensive language on several social media platforms. These standards are made after thorough research and feedback from the community. However, we are aware that the definitions in our guidelines are not representative of all possible perspectives. 
The degree of offensiveness scores that we provide in Ruddit are a representation of what the majority of our annotators think. We would like to emphasize that the scores provided are not the ``correct" or the only appropriate value of offensiveness. Different individuals and demographic groups may find the same comment to be more or less offensive than the scores provided.\\[-10pt]

\noindent {\bf Impact on Annotators:} Annotation of harsh and offensive language might impact the mental health of the annotators negatively \cite{vidgen-etal-2019-challenges, roberts2, roberts, ethics}. The following minimized negative mental impact on the annotators participating in our task: \\[-22pt]
\begin{itemize}
    \item The comments that we included in our dataset are pre-moderated by Reddit’s admins and subreddit specific moderators. Any comments that do not comply with Reddit’s content policy are not included.\footnote{\url{https://www.redditinc.com/policies/content-policy}} \vspace*{-3mm}
    \item Our goal was to annotate posts one sees on social media (after content moderation). Unlike some past work, we do not limit the data to include only negative comments. We included a large sample of posts that one normally sees on social media, and annotated it for degree of supportiveness or degree of offensiveness. \vspace*{-3mm}
    \item AMT provides a checkbox where requesters can indicate that some content in the task may be offensive. These tasks are not shown to annotators who have specified so in their profile. We used the checkbox to indicate that this task has offensive content.\vspace*{-3mm}
    \item We explicitly warned the annotators about the content of annotation, and advised worker discretion.\vspace*{-3mm}
    \item  We provided detailed annotation instructions and informed the annotators about how the annotations for offensive language will be used for studying and understanding offensive language.\vspace*{-3mm}
    \item The annotation of our data was crowdsourced, allowing for a large number of raters (725). This reduces the number of comments seen per rater. We also placed a limit on how many posts one may annotate. Annotators were not allowed to submit more than $\sim 5\%$ of the total assignments.\vspace*{-3mm}
    \item There are just 25 comments in the top 10\% of the offensiveness score range. Thus, most annotators ($>99.95\%$) do not see even one such comment.
\end{itemize}

\vspace*{-1mm}
\noindent {\bf Identity Terms:} As discussed in section \ref{sec:data-analysis}, in Ruddit, certain identity terms show a higher association with offensive comments than with the supportive comments. In order to address this, we created a variant of Ruddit, in which we replaced all the identity terms (from the list given in Appendix \ref{sec:identity}) with the [\textit{group}] token. We call this variant the \textit{identity-agnostic} dataset. 
We release the code for creating this variant from the original dataset. We evaluated our computational models on this variant and observed that the models did not learn to benefit from the association of the identity terms with the offensive comments.\\[-10pt]

\noindent {\bf Computational Models:} The models reported in this paper are not intended to fully automate offensive content moderation or to make judgements about specific individuals. Owing to privacy concerns, we do not model user history to predict offensiveness scores \citep{DBLP:journals/corr/abs-1810-03993}.\\[-10pt]

\noindent {\bf Feedback:} We are aware that our dataset is subject to the inherent bias of the data, the sampling procedure and the opinion of the annotators who annotated it. Finally, we acknowledge that this is not a  comprehensive listing of all the ethical considerations and limitations. We welcome feedback from the research community and anyone using our dataset.

\bibliographystyle{acl_natbib}
\bibliography{anthology,acl2021}

\clearpage
\appendix

\section{Supplemental Material}
\label{sec:appendix}

\subsection{Post and Comment Criteria}
\label{sec:postcriteria}
We selected the posts from the subreddits based on the following criteria:
\begin{enumerate}
    \item \textbf{Date}: To extract comments from posts that discuss current matters, we took comments from the time period of January, 2015 to September, 2019 (last available month at the time of extraction).
    \item \textbf{Thread length}: We chose posts with more than 150 comments and less than 5000 comments. This criteria ensured that the posts contained enough comments to capture meaningful discussion.
    \item \textbf{Post length}: We chose posts containing more than 5 words and less than 60 words in the post body. This was done to avoid posts that are too short to provide enough information or are too long and have a possibility of being spam. 
    \item \textbf{URL}: Often, posts on Reddit contain URLs redirecting to images, videos, news articles and others. We limited our posts to those containing at most one URL to avoid issues arising due to missing context.
\end{enumerate}

For each post, the hierarchical threads were reconstructed using the \emph{Anytree} python library.
We filtered comments from these posts based on the following criteria:
\begin{enumerate}
    \item \textbf{Comment length}: We chose comments containing more than 5 words and less than 150 words in the comment body. We did this to include comments that are neither too long (can be difficult to annotate) nor too short (not very valuable).
    \item \textbf{No. of users}: In the first and last 25 comments of the thread, we ensured participation of at least 4 users. This was done to ensure that the comments in our dataset are from a diverse set of users.
    \item \textbf{URL}: We chose comments with no URL in them. Comments with URL can be difficult to annotate as the URLs provide extra context for the comment. 
\end{enumerate}

\subsection{Annotation}
\label{sec:qa}

\begin{figure*}[t!]
\resizebox{\textwidth}{!}{%
  \includegraphics[]{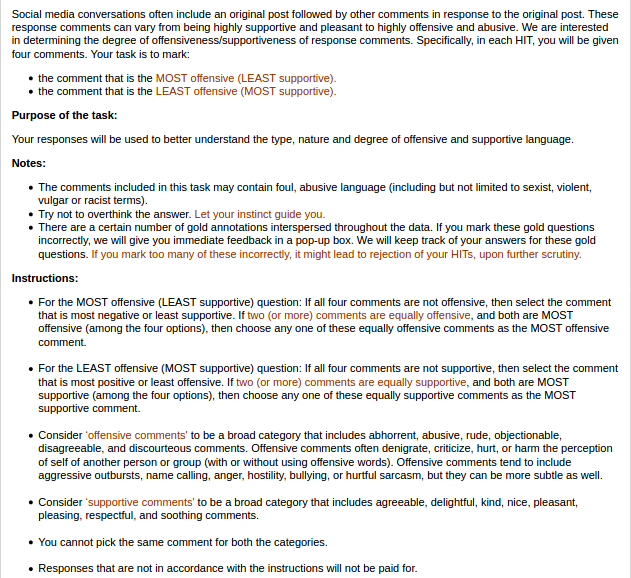}}
  \caption{Detailed instructions for the final annotation task.}
  \label{fig:instructions}
\end{figure*}

Figure \ref{fig:instructions} shows the detailed annotation instructions given to the crowd-workers for the task.

\begin{figure*}[t!]
\resizebox{\textwidth}{!}{%
  \includegraphics[]{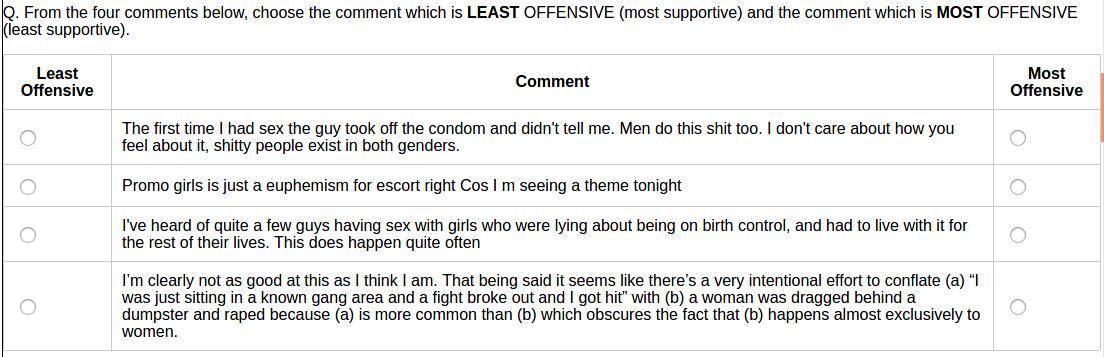}}
  \caption{Sample questionnaire for the final annotation task.}
  \label{fig:questionnaire}
\end{figure*}

A sample questionnaire for the final annotation task is shown in Figure \ref{fig:questionnaire}.

The hourly compensation rate for annotators on Amazon Mechanical Turk was US\$7.50/hr. 
The task received considerable attention with 725 participants in total.

\subsection{Sample data}
\label{sec:sampledata}
Table \ref{tab:comment-examples} contains comments from Ruddit grouped according to the 5 score bins. 

\begin{table*}[t!]
\centering
{\small
\begin{tabularx}{\textwidth}{lXl}
\toprule
\textbf{Bin} & \textbf{Comment}                                                                                                                 & \textbf{Score} \\ \midrule
1            & Truly gave me a smile                                 & $-0.812$       \\
             & Awwwwwe 
             That's was the best reveal I've seen.                                                                                            & $-0.688$        \\
             & I’m literally doing the same tonight!

Have an awesome time :)                                                                                             & $-0.625$       \\ 
2            & My apologies for missing that in your post.   Carry on.                                                                                            & $-0.583$       \\
             & I'm assuming this is a great response for i haven't seen the video in question, but i do believe that the woman in the video carried herself very elegantly!                           & $-0.375$        \\
             & Could you not love a ``purrfectly" healthy moggy? One whose life you could save from a shelter and won't break the bank.                                                                   & $-0.25$       \\ 
3            & Handwritten letters are pure gold. It is a simple but immensely pleasant gift to receive.  
Still have the ones my ex gf would sent me when we were apart.                                             & $-0.196$       \\
 &
  It would take way more time to get a screaming baby to stay still long enough to inject them. I remember my little sister throwing off doctors and nurses like a tiny she-hulk when she was a toddler. She also punched my dad in the face. &
  $-0.125$ \\
             & Oh man, that's a great idea! Nuke a few large cities in the US, this will send the economy to the sky! & $0.104$          \\ 
4            & No it’s usually their kids they are killing.      & $0.208$          \\
             & They were compassionately raped too. That's how you feel the love.
          & $0.312$            \\
             & \textbf{You look like a lesbian mechanic who has a shell collection}                               & $0.438$          \\ 
             & \textbf{You know it sucks when you cough, and the all of a sudden you got whooping gay}                                & $0.457$          \\  
             & \textbf{The best thing about anti-vax is soon we wont have to do anything about them, the diseases will do it for us}                                & $0.458$          \\  
5            & She's a hoe, plain and simple.                                                                            & $0.625$         \\
 &
  This guy is giving me fucking aids &
  $0.792$ \\
             & fuck you, you're just pretending to be blocked. get back to work RIGHT NOW you piece of shit piping     & $0.938$         \\ \bottomrule
\end{tabularx}%
}
\caption{More sample comments from Ruddit for each of the 5 score bins. Comments in bold are implicitly offensive.}
\label{tab:comment-examples}
\end{table*}

\subsection{Data Analysis}
\label{sec:identity}
We used the list of identity terms used by \citet{dixon_bias} with a few of our own additions. The terms used are \textit{lesbian, gay, bisexual, transgender, trans, queer, lgbt, lgbtq, homosexual, straight, heterosexual, male, female, nonbinary, african, africanamerican, black, white, european, hispanic, latino, latina, latinx, mexican, canadian, american, asian, indian, middle eastern, chinese, japanese, christian, muslim, jewish, buddhist, catholic, protestant, sikh, taoist, old, older, young, younger, teenage, millenial, middle aged, elderly, blind, deaf, paralyzed, atheist, feminist, islam, muslim, man, woman, boy, girl.
}

\begin{figure}[t!]
\centering
  \resizebox{\columnwidth}{!}{%
  \includegraphics[]{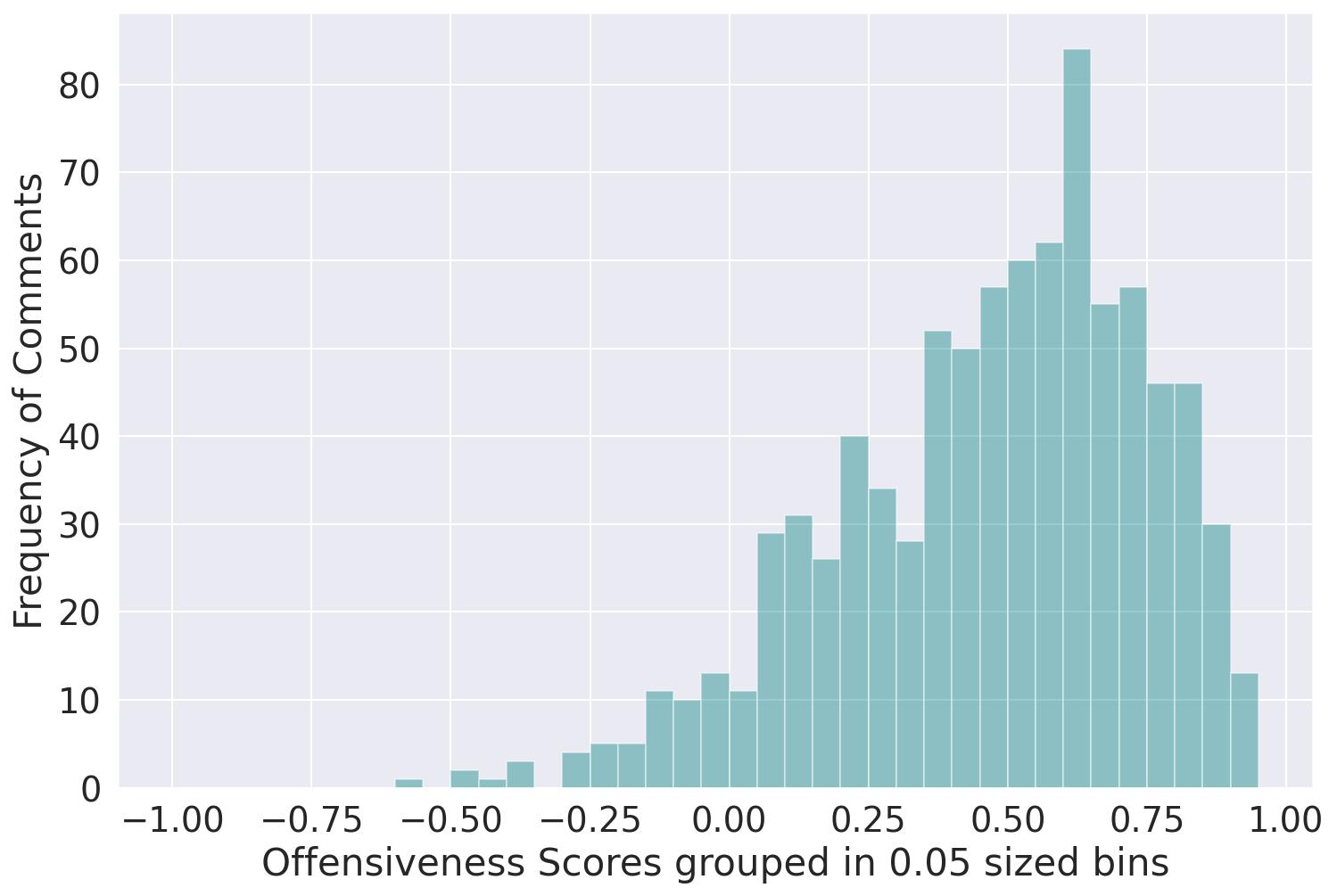}
  }
  \caption{A histogram of frequency of comments containing swear words--degree of offensiveness. Degree of offensiveness scores are grouped in bins of size $0.05$.}
  \label{fig:swear_dist}
\end{figure}

Figure \ref{fig:swear_dist} shows a histogram of the comments containing swear words--degree of offensiveness, over 40 equi-spaced score bins of size $0.05$.

\begin{figure*}[t!]
\centering
  \includegraphics[scale=0.35]{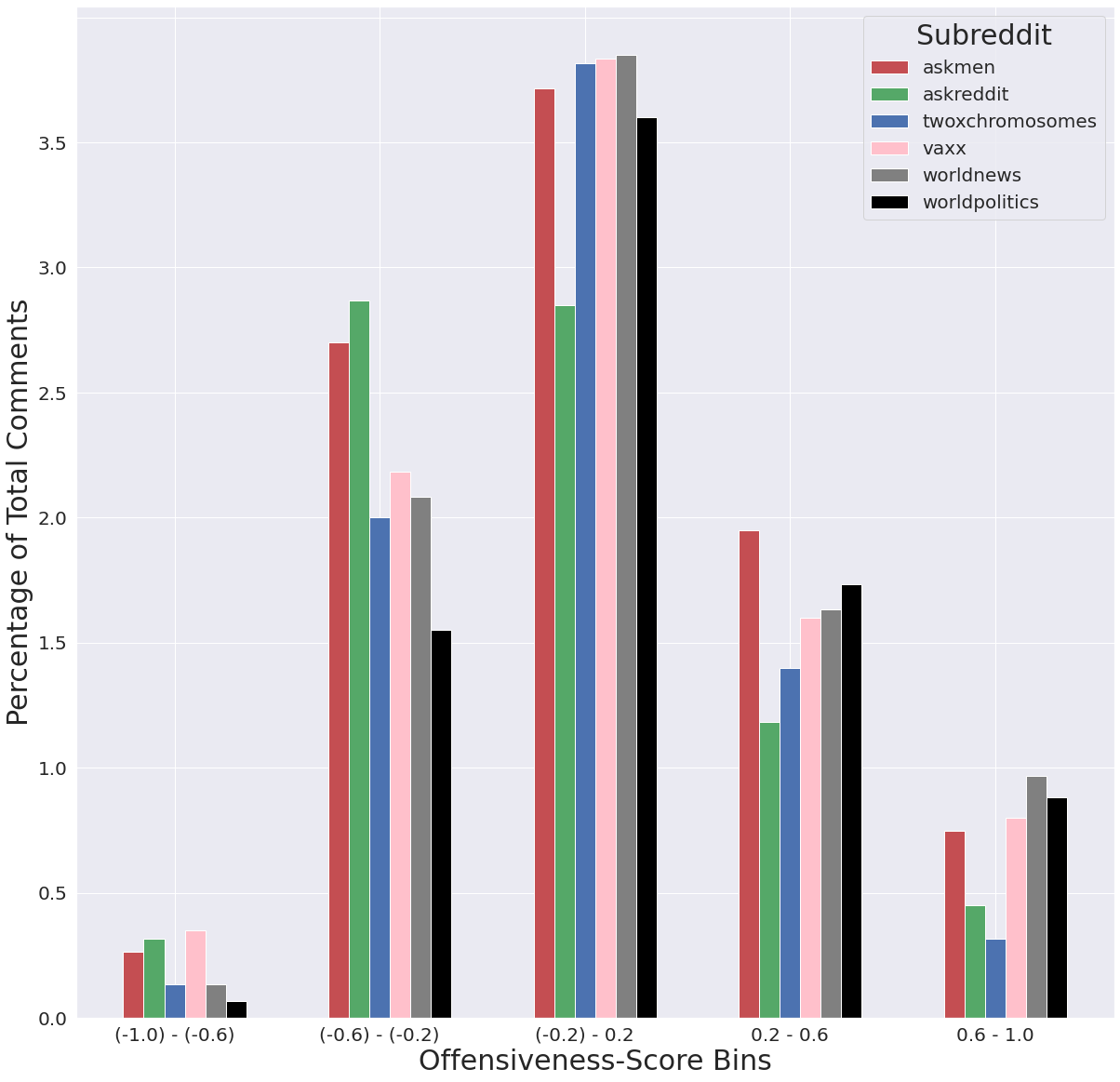}
  \caption{Distribution of comments from each subreddit in each offensiveness-score bin.}
  \label{fig:topic_dist}
\end{figure*}

Figure \ref{fig:topic_dist} shows a distribution of comments within each of the 5 score bins over the subreddits that were included in the Topics category.

\subsection{Computational Modeling}
\label{sec:gridsearch}
\paragraph{Hyperparameter Tuning} 
We tuned hyperparameters for the BERT and the BiLSTM models. We performed grid search cross-validation on Ruddit and used Pearson's $r$ to select the best hyperparameter setting. All experiments were performed on a fixed seed value of $12$.

For the BiLSTM model, the batch size was fixed at 32 and the number of epochs was set to 7. The hyperparameter search space is as follows: 

\begin{itemize}
    \item Number of Layers (N): $1, 2$
    \item Hidden size (H): $64, 128, 256$
\end{itemize}

For the BERT model, the batch size was fixed at 16 and BERT tokenizer's maximum length was set to 200. We tune hyperparameters on the settings that \citet{devlin-etal-2019-bert} found to work best on all tasks.  The search space is as follows:
\begin{itemize}
    \item Learning rate: $2e-5$, $3e-5$, $5e-5$
    \item Number of epochs: $3, 4$
\end{itemize}

We reported the best setting for the models in section \ref{sec:models}. The average $r$ of the BERT and the BiLSTM models across all hyperparameter search trials was $0.868 \pm 0.005$ and $0.827 \pm 0.002$ respectively. 

\paragraph{Training Times} We trained all our models on the Tesla T4 GPU. The number of GPU(s) used is 1. The number of trainable parameters and thus, the training time varied for each model. The approximate number of trainable parameters for each model is as follows: 
\begin{itemize}
    \item BiLSTM ($N=2$, $H=256$): $7$ million
    \item BERT: $108$ million
    \item HateBERT: $109$ million
\end{itemize}

\noindent The approximate average runtime for each model on the Ruddit dataset is as follows:
\begin{itemize}
    \item BiLSTM ($N=2$, $H=256$): $2$ seconds per epoch
    \item BERT: $3$ minutes per epoch
    \item HateBERT: $3.6$ minutes per epoch
\end{itemize}

\end{document}